\documentclass{article}

\usepackage[preprint]{neurips_2025}

\usepackage[utf8]{inputenc} 
\usepackage[T1]{fontenc}    
\usepackage{hyperref}       
\usepackage{url}            
\usepackage{booktabs}       
\usepackage{amsfonts}       
\usepackage{nicefrac}       
\usepackage{microtype}      
\usepackage{xcolor}         
\usepackage{graphicx}
\usepackage{multirow}
\usepackage{xcolor}
\usepackage{float} 
\usepackage{amsmath}
\usepackage{textcomp}
\usepackage{tcolorbox}
\usepackage{booktabs}
\usepackage{makecell}
\usepackage{arydshln}
\usepackage{wrapfig}
\usepackage{caption}
\title{Evaluating Mathematical Reasoning Across Large Language Models: A Fine-Grained Approach}

\author{%
  Afrar Jahin\thanks{Major contribution.$^{\dagger}$ Corresponding authors} \\
  School of Computer and Cyber Sciences \\
  Augusta University, Augusta, GA, USA \\
  \texttt{ajahin@augusta.edu} \\
  \And
  Arif Hassan Zidan\textsuperscript{$\dagger$} \\
  School of Computer and Cyber Sciences \\
  Augusta University, Augusta, GA, USA \\
  \texttt{azidan@augusta.edu} \\
  \And
  Wei Zhang\textsuperscript{$\dagger$} \\
  School of Computer and Cyber Sciences \\
  Augusta University, Augusta, GA, USA \\
  \texttt{wzhang2@augusta.edu} \\
  \And
  Yu Bao \\
  Department of Graduate Psychology \\
  James Madison University, Harrisonburg, VA, USA \\
  \texttt{bao2yx@jmu.edu} \\
  \And
  Tianming Liu \\
  School of Computing \\
  University of Georgia, Athens, GA, USA \\
  \texttt{tliu@uga.edu} \\
}

\begin{document}

\maketitle

\begin{abstract}
With the rapid advancement of Artificial Intelligence (AI), Large Language Models (LLMs) have significantly impacted a wide array of domains, including healthcare, engineering, science, education, and mathematical reasoning. Among these, mathematical reasoning remains a particularly challenging capability, often requiring multi-step logic and abstract generalization. While prior work has explored LLM performance on reasoning tasks, comprehensive evaluations that span both depth and breadth across model families remain limited. In this study, we present a systematic evaluation of mathematical reasoning abilities across eight leading LLMs, including two recent DeepSeek models, using three independent benchmark datasets. Our analyses reveal several key findings: (1) DeepSeek-R1 performs competitively with o1 across most domains and achieves the highest accuracy on the MMLU Formal Logic benchmark; (2) distilled variants, such as DeepSeek-1.5B, exhibit substantial performance degradation; and (3) Gemini 2.0 Flash achieves the lowest response latency. Beyond quantitative metrics, we explore how architectural choices, training paradigms, and optimization strategies contribute to variation in reasoning performance. These findings provide new insights into the capabilities and limitations of current LLMs in mathematical domains, and offer guidance for the development of future models better aligned with rigorous reasoning demands.

\end{abstract}

\section{Introduction}

As Artificial Intelligence (AI) continues to advance at an unprecedented pace, an array of powerful Large Language Models (LLMs) has emerged, including OpenAI’s GPT-4o, o1, o3, Claude 3.5, Llama 4, Qwen 3, Gemini 2.0 and so on~\citep{achiam2023gpt,chang2024survey,dubey2024llama,enis2024llm,kasneci2023chatgpt,liu2024deepseek,team2024gemma,zhao2023survey}. Driven by cutting-edge advancements in Deep Neural Networks (DNNs), these models integrate human cognition elements to enhance problem-solving and decision-making~\citep{lappin2024assessing}. Serving as a guiding light in natural language processing (NLP)~\citep{min2023recent}, healthcare~\citep{dai2023ad,liu2023holistic,liu2023radiology}, clinical textual processing~\citep{dai2025auggpt,liu2023deid,zhong2023chatabl}, biomedical image analyses~\citep{liu2023holistic}, code generation~\citep{liu2023your}, decision support~\citep{li2022pre}, multimodal data analytics~\citep{wang2024comprehensive,xiao2024instruction}, and mathematical reasoning~\citep{ahn2024large,li2024evaluating,peng2024multimath,shao2024deepseekmath,xin2024deepseek,zhong2024evaluation,zhuang2024math}, LLMs push the boundaries of AI capabilities, approximating human-like reasoning through sophisticated statistical inference~\citep{achiam2023gpt,kasneci2023chatgpt,neha2025survey}. However, despite their transformative potential, these models face notable limitations. Their high computational demands pose significant barriers to broader accessibility, making large-scale implementation costly~\citep{neha2025survey}. Moreover, while LLMs perform well in general contexts, they often struggle with specialized tasks, exhibiting inconsistencies in performance. Multimodal models, for example, continue to face challenges in spatial reasoning and real-world physics, while AI-assisted code generation frequently produces syntactically correct yet functionally flawed outputs, requiring human oversight~\citep{liu2023your,zhong2024evaluation}. These constraints underscore the ongoing need for refinement and innovation in AI research to bridge the gap between artificial and human intelligence.

In 2024, OpenAI released GPT-4o, a multimodal model capable of processing text, images, and audio. With estimated 2 trillion parameters, it significantly surpasses GPT-3 in tasks such as reasoning, coding, and clinical diagnostics~\citep{achiam2023gpt,dale2021gpt,hurst2024gpt,shahriar2024putting}. Meanwhile, other cutting-edge models, such as o1, o3 and recently o4-mini, have been introduced between 2024 and 2025. In particular, o1 enhances reasoning capabilities by incorporating a Chain-of-Thought (CoT) approach~\citep{stechly2025chain, zhong2024evaluation}, where it renders intermediate reasoning steps before reaching a final answer closely mirroring the way humans process complex problems~\citep{arrieta2025early}.

In addition, Claude 3.5, released in 2024, emphasizes safety, alignment, and performance, with improvements in reasoning, language understanding, and handling complex tasks like text and code generation~\citep{enis2024llm}. With 250B parameters, it surpasses earlier models in accuracy and ethical alignment. It supports up to 200K tokens for extended context. Notably, enhanced by reinforcement learning from human feedback (RLHF)~\citep{hu2024dawn,wen2024language} and Constitutional AI, it reduces undesirable responses, biases, and better aligns with human intent. Claude 3.5 excels in specialized areas like coding and scientific reasoning, with improved transparency and ethical safeguards~\citep{enis2024llm,neha2025survey}. 

Furthermore, Llama-3.3, presented in 2024, is the latest version of LLM in Meta AI family, following Llama-1 and Llama-2. Llama-3.3 advances further with 70B parameters and 128K token context window, improved by grouped-query attention for better efficiency~\citep{dubey2024llama,lu2024small}. Llama 3.1 excels in coding, logical problem solving, and low-resource language tasks. Unlike closed models such as GPT series, it remains open-weight and freely accessible for research and commercial use, but is restricted to text-only input~\citep{dominguez2025questioning,lu2024small,neha2025survey}. 

Moreover, Gemini 2.0 Flash, initially introduced as an experimental variant, provides significant speed and efficiency gains over its predecessor, Gemini 1.5 Flash, without sacrificing the efficiency~\citep{huang2024survey,imran2024google,team2024gemma}. Notably, it outperforms Gemini 1.5 Pro on key benchmarks while operating at twice the speed. It enables the incorporation of agentic AI and native use, allowing the model to call external functions (Google Search and Maps) and integrate streaming data for expanded real-time applications. By combining better performance in tasks such as math, code generation, and multilingual audio output with enhanced efficiency. Gemini 2.0 aims to deliver comprehensive, cost-effective AI solutions for both developers and end users~\citep{huang2024survey,team2024gemma}. Lastly, Qwen2.5, released in September 2024, is the latest iteration in the Qwen series, following Qwen2 in June 2024 and the original Qwen in August 2023. Qwen1.5 featured models up to 72B parameters, emphasizing efficiency and open-source accessibility. Qwen2 introduced improved reasoning, multilingual support, and coding capabilities, with models scaling up to 72.71B parameters~\citep{yang2024qwen2, zhu2025qwen}. 

Besides, established in 2023 as a research initiative to push the boundaries of artificial general intelligence (AGI), DeepSeek models set out to overcome existing limitations by developing specialized models focused on efficiency, adaptability, and domain expertise~\citep{bi2024deepseek,dai2024deepseekmoe,guo2024deepseek,guo2025deepseek,liu2024deepseek,kasneci2023chatgpt,liu2024deepseekv2,shao2024deepseekmath}. In 2024, the Mixture-of-Experts (MoE), an efficiency-driven architecture that leverages sparse activation to reduce computational overhead was introduced~\citep{guo2024deepseek,liu2024deepseekv2,liu2024deepseek}. Meanwhile, DeepSeek Math, trained on 120B math-related tokens, was developed to handle advanced mathematical and symbolic reasoning tasks~\citep{shao2024deepseekmath}. In 2025, DeepSeek made a significant breakthrough with R1 Zero, incorporating self-verification, reflection, and extended CoTs. Notably, DeepSeek-R1 was recently presented, specifically designed for mathematical, coding, and logical problem-solving, enhancing autonomous decision-making and precision in both research and enterprise applications~\citep{guo2025deepseek,guo2024deepseek,shao2024deepseekmath}. 

Notably, mathematical reasoning poses intricate challenges without straightforward solutions~\citep{ahn2024largelanguagemodelsmathematical,liang2024mathematics}. Unlike routine tasks that follow established frameworks, mathematical reasoning demands creativity, abstract thinking, and advanced cognitive skills. In this work, we systematically evaluate the mathematical reasoning performance of a wide array of LLMs, with a particular emphasis on DeepSeek models, to further unveil machine creativity in mathematical reasoning. Leveraging multiple independent datasets and quantitative metrics, we conduct a comprehensive empirical analysis to assess and compare their capabilities. Our findings render a detailed evaluation of prominent LLMs in mathematical reasoning while highlighting the advancements and unique strengths of LLMs.

\section{Experimental Framework}
\label{sec2}

In this section, we provide a comprehensive evaluation of our experimental framework, detailing the datasets utilized, the quantitative evaluation metrics, as well as the overall methodology. We also present a justification for selecting these datasets as benchmarks, emphasizing their relevance, diversity, and suitability in effectively assessing LLM performance in mathematical reasoning. By establishing a consistent evaluation framework, we aim to ensure a thorough and insightful comparison of model capabilities across various reasoning tasks.

\begin{figure}[h]
    \centering
    \includegraphics[width=1.00\linewidth]{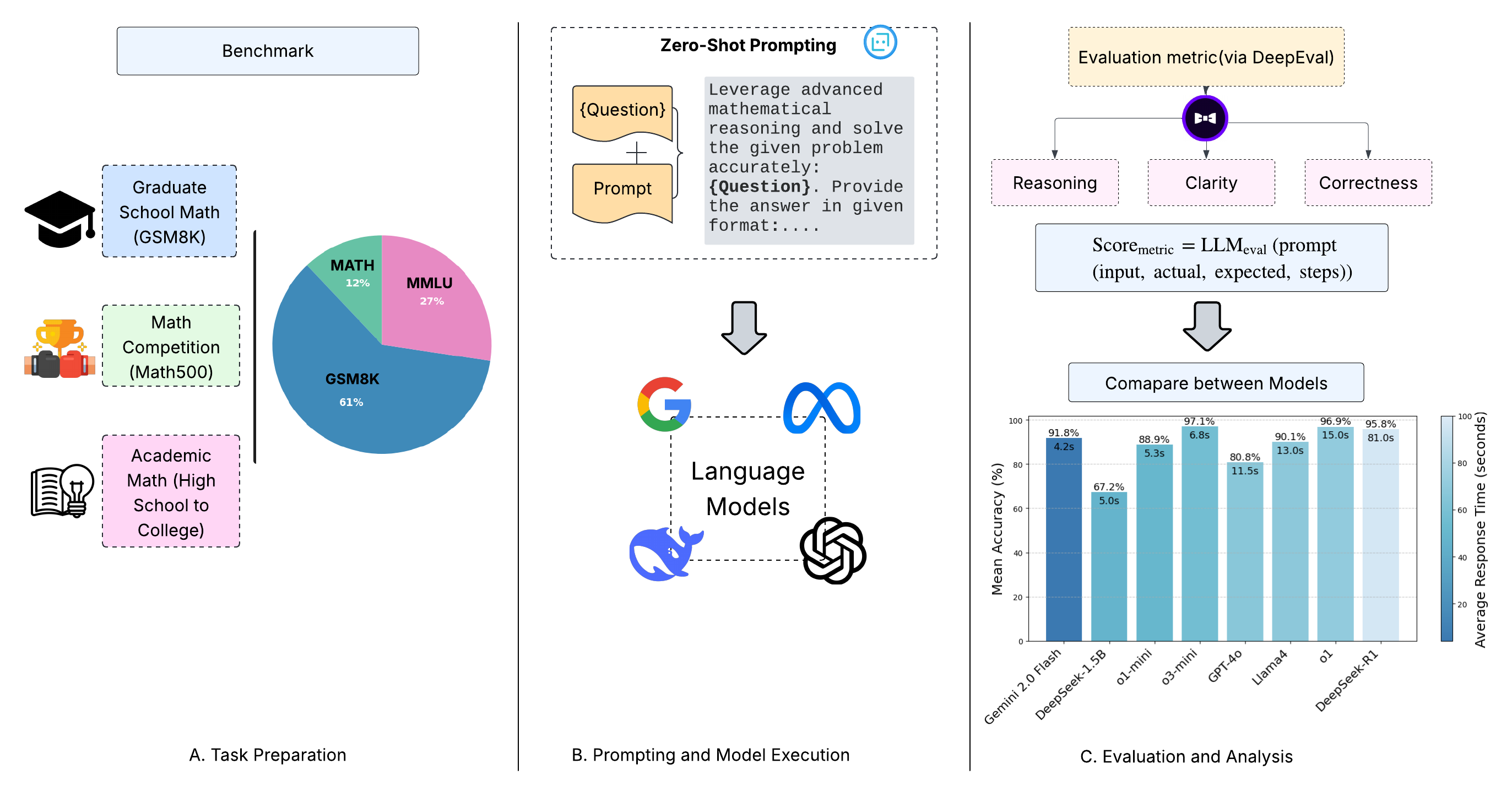} 
    \caption{Overview of the experimental framework: (A) Task preparation involves selecting benchmark datasets spanning academic, graduate-level, and competition math problems. (B) Zero-shot prompting is used to query various LLMs with structured instructions. (C) Model outputs are evaluated using DeepEval across reasoning, clarity, and correctness dimensions, enabling quantitative comparison of model performance.}
    \label{fig:experimental_framework}
\end{figure}

\subsection{Experimental Designs}
\label{subsec2.1}

The framework of this empirical study is illustrated in Figure~\ref{fig:experimental_framework} and consists of three sections: A) We selected benchmarks consisting of diverse math datasets.~\citep{cobbe2021gsm8k,hendrycks2020measuring,lu2023mathvista} B). Zero-Shot Prompting: Zero-shot prompting was employed to design all prompts~\citep{hu2025localized}, ensuring a standardized evaluation approach. The prompts are uniformly applied across a suite of language models.~\citep{imran2024google,achiam2023gpt, guo2025deepseek, arrieta2025early, zhong2024evaluation, liu2024deepseekv2}(C) We evaluate model outputs using DeepEval's \citep{Ip_deepeval_2025} fine-grained metrics: correctness, clarity, and reasoning, each computed by comparing predicted answers to ground truth under structured evaluation steps. Overall, this structured framework enabled holistic and systematic evaluations of LLMs in mathematical reasoning.

\subsection{Datasets}
\label{subsec2.2}
To evaluate the mathematical reasoning capabilities of LLMs, we employ well-established benchmark datasets that encompass diverse mathematical domains, question types, and difficulty levels. The selected datasets include Math Competition (MATH)~\citep{hendrycksmath2021}, Grade School Math (GSM8K)~\citep{cobbe2021gsm8k}, and Massive Multitask Language Understanding (MMLU)~\citep{hendrycks2020measuring} math subset, covering topics ranging from elementary arithmetic to advanced algebra, logic, and mathematical competitions. From each benchmark, we carefully select a representative subset of problems to ensure a balanced evaluation while maintaining the diversity and complexity of mathematical reasoning tasks. These subsets are chosen for their broad coverage of mathematical domains, variety in question formats, and representation of different complexity levels, ensuring a comprehensive assessment of LLMs' symbolic manipulation, logical reasoning, and problem-solving abilities~\citep{gema2024we,setlur2024rl,zhang2024careful}. 

\begin{wrapfigure}{r}{0.5\textwidth}
    \centering
    \vspace{-10pt} 
    \includegraphics[width=\linewidth]{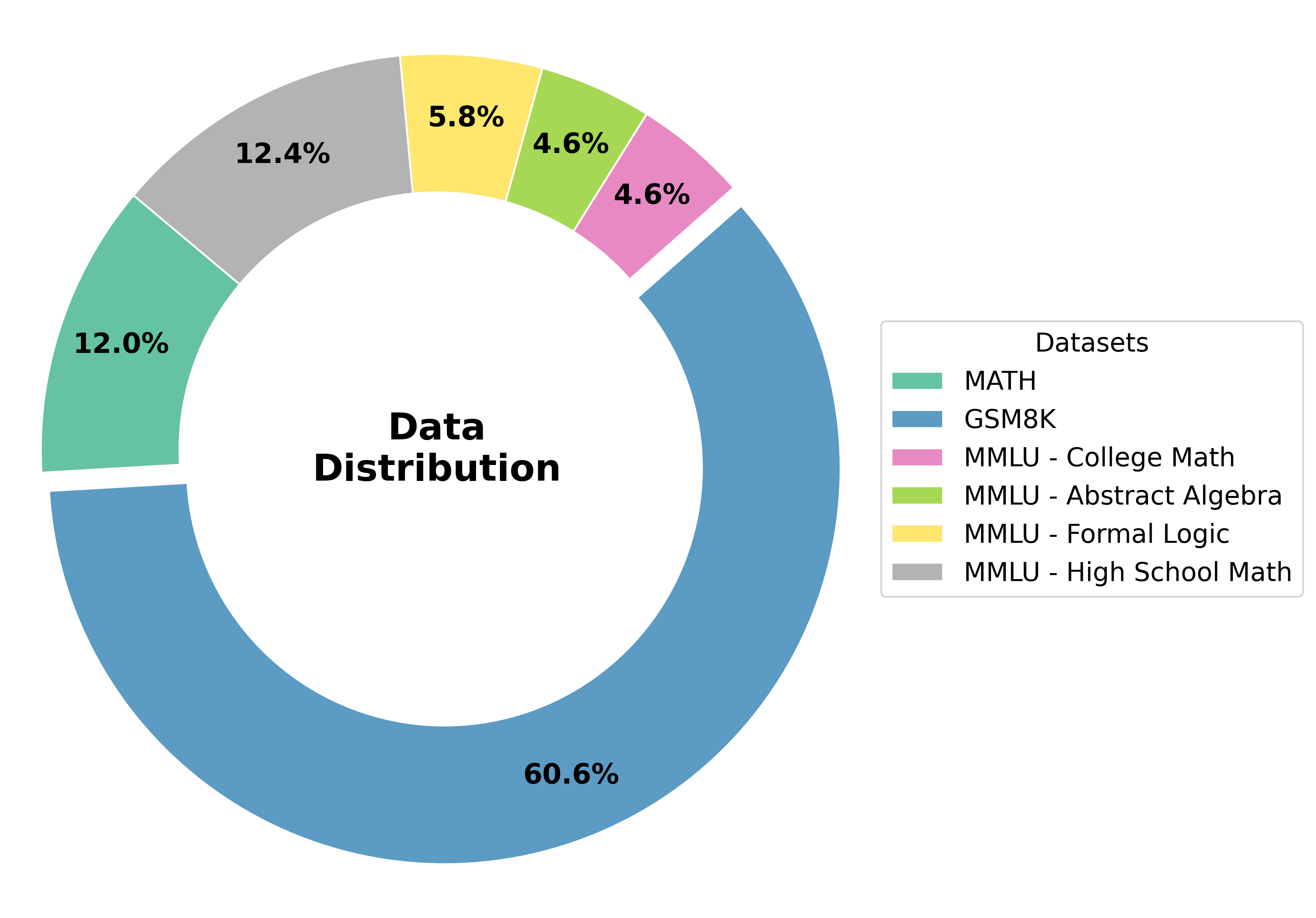}
    \caption{Data distribution across selected datasets for evaluating mathematical reasoning.}
    \label{fig:data_distribution}
    \vspace{-10pt} 
\end{wrapfigure}

Notably, these subsets are chosen for their broad coverage of mathematical domains, variety in question formats, and representation of different complexity levels, ensuring a comprehensive assessment of LLMs' logical reasoning, and problem-solving capabilities~\citep{gema2024we,setlur2024rl,zhang2024careful}. We selected 262 free-form numeric problems from \textbf{MATH} dataset (mathematics competitions), covering domains such as algebra, geometry, number theory, and combinatorics. For this study, we selected problems spanning \(\geq 3\) difficulty levels to ensure coverage between medium and hard tasks. The \textbf{GSM8K} testset includes 1,320 arithmetic word problems designed for the grade school level. From \textbf{MMLU}, we incorporate four subject-specific subsets—\textit{College Mathematics (MCM)}, \textit{Abstract Algebra (MAA)}, \textit{Formal Logic (MFL)}, and \textit{High School Mathematics (MHM)}—ranging from 100 to 270 multiple-choice questions. These subsets cover diverse topics such as calculus, group theory, logical reasoning, and general high school math. In total, our experimental benchmark includes 2,178 problems, offering broad coverage across difficulty levels, formats, and mathematical domains to support fine-grained evaluation of reasoning skills in LLMs. By evaluating LLMs on these benchmarks, we identified their strengths and limitations in mathematical reasoning, aiding in the development of more robust and capable models
Figure~\ref{fig:data_distribution} presents an overview structure of the datasets utilized in this study. The detailed table of the dataset is shown in Appendix A.

\subsection{Evaluation of Mathematical Reasoning Across LLMs}
\label{subsec2.3}
As discussed previously, multiple LLMs have demonstrated remarkable capabilities in mathematical reasoning~\citep{zhou2024your}. To systematically and comprehensively evaluate the mathematical reasoning capabilities of various LLMs, we conducted experiments using API-based access to a diverse set of models. The selected models include  OpenAI’s GPT-4o, o1-mini, o1, and o3-mini, DeepSeek’s DeepSeek-1.5B and DeepSeek-R1, Google’s Gemini 2.0 Flash, and Meta's Llama-4 the latest one. These models were accessed through their respective API services using secure API keys from Google, OpenAI, DeepSeek, and Openrouter.  

\subsubsection{Prompting Strategy} 
\label{subsec2.3.1}
The effectiveness of LLMs in mathematical reasoning is highly dependent on the design of input prompts~\citep{amatriain2024prompt,cain2024prompting,denny2024prompt}. To ensure a fair and reproducible evaluation, we employed a structured prompting strategy focusing on zero-shot prompting initially. Our methodology was designed to assess models' inherent problem-solving abilities without additional external guidance~\citep{wu2024get}. Overall, for all evaluations, we incorporated a zero-shot prompting approach, where models were presented with mathematical problems without any prior exemplars or demonstrations. Zero-shot prompting is particularly valuable in assessing a model’s ability to generalize mathematical knowledge from its pretraining corpus and apply learned concepts to novel problems~\citep{wu2024get,yuan2024instance}. By not providing explicit examples, this method evaluates how well a model can infer the appropriate solution methodology based solely on its prior learning.

For multiple-choice questions, we provided the set of answer choices and instructed the models to select the correct response from the given options, ensuring a structured decision-making process~\citep{myrzakhan2024open}. In contrast, for open-ended problems that required free-form numeric or symbolic responses, we presented the question directly without predefined answer choices, allowing the models to generate responses based on their internal reasoning capabilities~\citep{tam2025none}. This approach ensured that multiple-choice tasks evaluated the models' ability to differentiate between structured options. The prompt templates linked with each category question are elaborated in Appendix B.

\subsubsection{Settings}  
\label{subsec2.3.2}
To ensure a standardized evaluation across different LLMs, we utilized API access through OpenAI and OpenRouter, while maintaining their respective default generation settings. All models, including GPT-4o and o-series models, Gemini, Llama-4 and DeepSeek use a default temperature of $T=1.0$ and an unrestricted probability sampling strategy (top-p = 1.0)~\citep{nguyen2024turning}, allowing the model to consider the full probability distribution for token selection. 

\subsubsection{Evaluation Metric}
\label{subsec2.3.3}
We evaluate the model outputs using three core metrics, \textit{Correctness}, \textit{Clarity}, and \textit{Reasoning}, which reflect different dimensions of mathematical understanding and communication. The evaluation is conducted using \textit{LLM-as-a-judge} methodology implemented via the DeepEval framework.\citep{Ip_deepeval_2025}

\textbf{Correctness} measures whether the model output reaches the right final answer or conveys the correct mathematical content, allowing for variations in formatting. \textbf{Clarity} assesses the logical coherence and readability of the explanation, ensuring that the reasoning process is easy to follow. \textbf{Reasoning} evaluates the depth and validity of the model’s step-by-step problem-solving approach.

Each model-generated response is processed through a pre-defined prompt template with access to the original question and reference answer. DeepEval internally prompts a trusted evaluator model (e.g., GPT-4o) to return scalar ratings for each metric. Let \( M \) be a model, \( Q \) be a set of questions, and \( \text{Score}_{M}^{(k)}(q) \) be the score for metric \( k \in \{\text{Correctness}, \text{Clarity}, \text{Reasoning}\} \) on question \( q \in Q \). The average metric score is computed as:

\[
\text{Score}_{M}^{(k)} = \frac{1}{|Q|} \sum_{q \in Q} \text{LLM}_{\text{eval}}\left( \text{prompt}(q, \text{output}_M(q), \text{gold}(q), \text{steps}_k) \right)
\]

\section{Results and Analysis}
\label{sec3}
In this section, we conducted a quantitative and qualitative evaluation of the state-of-the-art LLMs mentioned earlier for mathematical reasoning. As an initial assessment, we selected a subset of three widely recognized datasets~\citep{cobbe2021gsm8k,hendrycks2020measuring,lu2023mathvista}, as introduced in Section~\ref{sec3}, to systematically evaluate the performance of various cutting-edge models. Overall, our evaluation in Table~\ref{tab:evaluation_metric_result}, covering diverse evaluation metric, such as Accuracy(Acc), Reasoning Score(RS), and Clarity Score(CS), includes DeepSeek-R1~\citep{liu2024deepseekv2},along with its distilled variant DeepSeek-1.5B~\citep{guo2025deepseek}, which is derived from the R1 model. Additionally, we assessed Google’s latest Gemini 2.0 Flash~\citep{imran2024google}, four OpenAI models: GPT-4o, o1-mini, o1, and o3-mini~\citep{arrieta2025early,hurst2024gpt,zhong2024evaluation} and finally Meta's Llama-4 the latest model~\citep{meta2025llama4}. Notably, while all these models are designed with reasoning capabilities, GPT-4o and Llama-4 are not explicitly optimized for providing reasoning-based CoT responses but they can still perform CoT reasoning quite well when prompted, especially in instruction-tuned variants. These selected LLMs represent the most advanced AI systems available today, continuously evolving and competing across various mathematical reasoning benchmarks.

\begin{table*}[t]
\centering
\small
\setlength{\tabcolsep}{4pt}
\renewcommand{\arraystretch}{1.2}
\caption{Performance comparison of LLMs across multiple mathematical reasoning abilities using three metrics: Accuracy (ACC), Reasoning Score (RS), and Clarity Score (CS). The results highlight the average performance of each model across all categories.}
\label{tab:evaluation_metric_result}
\begin{tabular}{|l|cccccc|c|}
\hline
\textbf{LLMs} & \textbf{Math} & \textbf{GSM8K} & \textbf{MAA} & \textbf{MFL} & \textbf{MCM} & \textbf{MHM} & \textbf{Avg.} \\
\hdashline
\multicolumn{8}{|c|}{\textbf{\textcolor{purple}{Accuracy – $ACC$ (\%)}}} \\
\hdashline
GPT-4o           & 67.56 & 96.21 & 86.00 & 82.54 & 79.00 & 88.89 & 80.80 \\
Deepseek-R1      & 92.37 & 96.36 & 95.00 & \textbf{97.62} & 96.00 & 97.78 & 95.75 \\
Deepseek-1.5B    & 68.70 & 81.58 & 61.00 & 52.38 & 66.00 & 88.15 & 67.25 \\
o1               & 95.80 & 96.21 & 94.00 & 96.51 & \textbf{99.00} & \textbf{99.26} & 96.91 \\
o1-mini          & 89.69 & 96.13 & 88.00 & 78.57 & 89.29 & 98.89 & 88.89 \\
o3-mini          & \textbf{96.56} & 94.84 & \textbf{96.00} & 95.24 & \textbf{99.00} & 98.52 & \textbf{97.06} \\
Gemini-2.0-Flash & 87.02 & 94.24 & 88.00 & 90.48 & 96.00 & 97.77 & 91.85 \\
Llama-4          & 81.68 & \textbf{97.04} & 88.00 & 90.48 & 93.00 & 97.41 & 90.11 \\
\hdashline
\multicolumn{8}{|c|}{\textbf{\textcolor{purple}{Reasoning Score – $RS$ (\%)}}} \\
\hdashline
GPT-4o           & 90.41 & \textbf{98.93} & \textbf{98.60} & 94.72 & 95.20 & 94.69 & 94.32 \\
Deepseek-R1      & 95.15 & 98.15 & 94.76 & 94.04 & 98.26 & 98.19 & 96.08 \\
Deepseek-1.5B    & 77.58 & 88.56 & 72.54 & 66.95 & 80.82 & 89.33 & 77.64 \\
o1               & 98.73 & 98.82 & 94.00 & 86.84 & \textbf{99.09} & 98.20 & 95.97 \\
o1-mini          & 95.51 & 98.46 & 87.49 & 95.63 & 93.09 & \textbf{98.51} & 94.04 \\
o3-mini          & \textbf{99.04} & 97.53 & 97.37 & \textbf{98.44} & 99.05 & 98.33 & \textbf{98.45} \\
Gemini-2.0-Flash & 89.52 & 96.01 & 93.54 & 93.78 & 92.07 & 97.12 & 93.60 \\
Llama-4          & 92.34 & 98.15 & 93.52 & 95.59 & 95.54 & 97.99 & 94.59 \\
\hdashline
\multicolumn{8}{|c|}{\textbf{\textcolor{purple}{Clarity Score – $CS$ (\%)}}} \\
\hdashline
GPT-4o           & 83.58 & 98.32 & 89.54 & 88.77 & 87.22 & 86.52 & 87.33 \\
Deepseek-R1      & 85.89 & 97.21 & 87.81 & 89.13 & 84.42 & 88.10 & 87.07 \\
Deepseek-1.5B    & 77.21 & 88.33 & 60.09 & 66.27 & 80.27 & 83.27 & 73.62 \\
o1               & 86.80 & \textbf{98.34} & 84.91 & 77.76 & 85.00 & 87.95 & 84.48 \\
o1-mini          & \textbf{87.87} & 98.11 & 80.52 & 89.64 & 81.29 & 88.25 & 85.11 \\
o3-mini          & 87.43 & 96.88 & \textbf{89.89} &  \textbf{91.55} & \textbf{91.26} & 88.26 & \textbf{89.07} \\
Gemini-2.0-Flash & 80.41 & 90.48 & 81.93 & 88.13 & 79.98 & 85.23 & 83.14 \\
Llama-4          & 87.73 & 94.71 & 81.79 & 89.85 & 85.69 & \textbf{89.07} & 86.43 \\
\hline
\end{tabular}
\end{table*}

\subsection{Overview of Benchmark Performance}
\label{subsec3.1}
We begin our evaluation with a comprehensive comparison of overall accuracy across the benchmark tasks, including MATH, GSM8K, and four mathematical subsets from the MMLU corpus(MAA, MFL, MCM, MHM). Table~\ref{tab:evaluation_metric_result} summarizes the accuracy score (\textbf{ACC}), clarity score (\textbf{CS})and reasoning scores (\textbf{RS}), highlighting each model's performance and robustness across these domains.

Overall, the o3-mini model achieved the highest average accuracy, outperforming all baselines with an average of 97.06\%. It showed particularly strong performance in the Math (96.56\%), MAA (96.00\%), and MCM (99.00\%) subsets. Similarly, DeepSeek-R1 and o1 demonstrated competitive accuracy, achieving 95.75\% and 96.91\% respectively. When evaluating reasoning score, o3-mini again led the comparison with an average score of 98.45\%, closely followed by DeepSeek-R1 (96.08\%) and o1 (95.97\%). These models consistently showed high reasoning coherence and correctness across all mathematical domains. Importantly, we include representative examples of both correct and incorrect responses in Appendix C.

In terms of clarity score, which evaluates the explanation's readability and logical structure, the results followed a similar trend. o1 and o1-mini scored highest with averages of 88.44\% and 85.11\%, respectively. Notably, GPT-4o, although not optimized explicitly for math reasoning, maintained competitive scores in clarity (87.33\%) and reasoning (94.32\%).


Notably, the distilled variant DeepSeek-1.5B exhibited markedly lower scores across all three metrics, especially in reasoning (77.64\%) and clarity (73.62\%), emphasizing the performance compromises introduced by model compression. Appendix D provides a detailed comparative breakdown of each model’s performance.

In summary, while all evaluated models show proficiency, our findings reveal significant variance in cross-domain generalization. The superior performance of models like DeepSeek-R1 and o1 across diverse mathematical tasks positions them as current leaders in LLM-based mathematical reasoning, while others specially deepseek's distilled variant, despite it's scale, remain constrained by limitations in symbolic abstraction and formal manipulation. 

\subsection{Cognition Profile Across Reasoning Dimensions}
\label{subsec3.2}
\begin{figure}[t]
    \centering
    \includegraphics[width=1.0\linewidth]{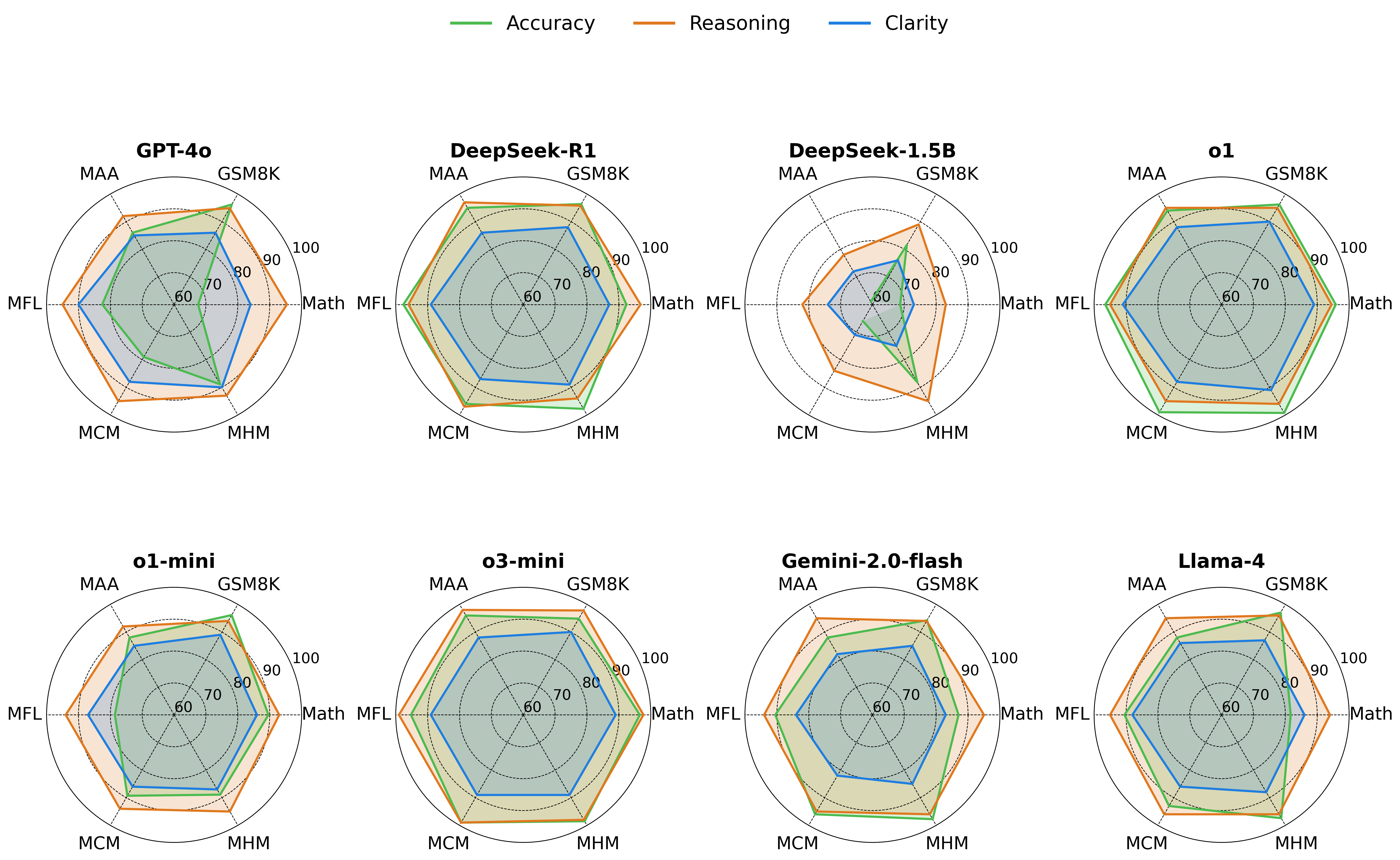} 
    \caption{Radar plots of LLM performance across six mathematical domains. Each sub-plot corresponds to a model. Lines represent three evaluation metrics: Accuracy, Reasoning, and Clarity. This visualization highlights the strengths and weaknesses of each model across different reasoning dimensions and content types. }
    \label{fig:radar_plot}
\end{figure}

To better understand the model-wise variation in mathematical reasoning behavior, we visualize the cognition profiles of each evaluated LLM using radar plots (Figure~\ref{fig:radar_plot}). Each subplot corresponds to a single model, showing its relative performance across six domains: MATH, GSM8K, MAA, MFL, MCM, and MHM. The colored lines represent the evaluation dimensions—Accuracy, Reasoning, and Clarity—as measured by the DeepEval framework~\citep{Ip_deepeval_2025}. The Figure~\ref{fig:radar_plot} demonstrates, o3-mini demonstrates consistent and high performance across all dimensions, with particularly dominant accuracy and reasoning scores, forming a near-regular and saturated polygon. This consistency highlights the effectiveness of its architecture in capturing logical structure and factual correctness.

In contrast, the DeepSeek-1.5B (distilled) model displays notable asymmetry in its radar shape, indicating uneven performance across both domains and evaluation metrics. Specifically, the reduced reasoning and clarity scores in the MAA and MFL domains suggest limitations in handling abstract and symbolic reasoning, a likely result of aggressive distillation. On the other hand, GPT-4o model shows strong reasoning capability, especially in GSM8K and MFL, though with relatively lower accuracy in the MATH domain. Its clarity curve remains stable, affirming its general-purpose alignment rather than domain-specific tuning. DeepSeek-R1 and o1 both display high and balanced performance, especially in the reasoning dimension, validating their suitability for high-precision mathematical applications. Meanwhile, Llama-4 and Gemini 2.0 Flash show strong clarity but slightly lower reasoning compared to the top-tier models.

Overall, these cognition profiles emphasize that high performance in mathematical reasoning tasks depends not just on accuracy but also on the model's ability to maintain logical flow and human-readable clarity across domains. The radar plots provide an interpretable and multi-dimensional view into how different models prioritize or trade off among these cognitive capabilities.


\subsection{Accuracy--Latency Trade-off Analysis}
\label{subsec3.3}
\begin{wrapfigure}{r}{0.7\textwidth}
    \centering
    \includegraphics[width=0.8\linewidth]{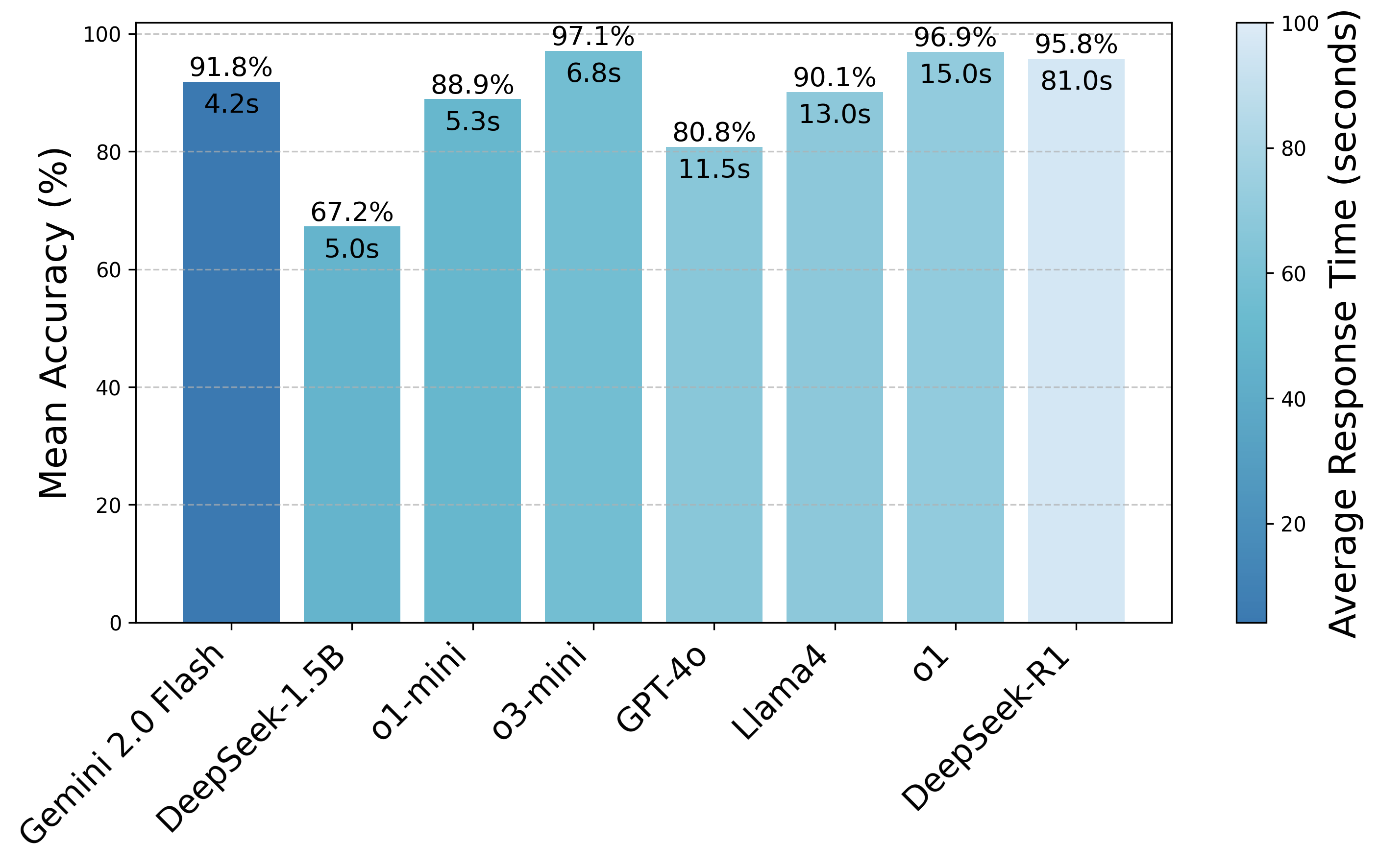} 
    \caption{Mean accuracy (\%) versus average response time (seconds) for each model. The bar height indicates average accuracy across reasoning tasks, while the color gradient and labels represent the average response latency. }
    \label{fig:res_acc_vs_response_time}
\end{wrapfigure}

Efficient response times are crucial in the deployment of LLMs, as they directly influence user experience and the practicality of real-time applications. Factors affecting inference speed include model size, computational complexity, hardware capabilities, and optimization techniques.

To complement our evaluation of cognitive performance, we investigate the trade-off between reasoning accuracy and computational efficiency across models (Figure~\ref{fig:res_acc_vs_response_time}). Each bar represents a model's average accuracy across our mathematical reasoning benchmarks, while the color gradient and numerical labels denote the average response time (in seconds), measured using consistent inference settings.

Among the models, o3-mini achieves an optimal balance, with a high accuracy of 97.1\% and a relatively low response time of 6.8 seconds. This performance places it at the efficiency frontier, demonstrating both computational effectiveness and strong reasoning capabilities. Similarly, o1-mini maintains respectable accuracy (88.9\%) with an even lower latency of 5.3 seconds, making it highly suitable for real-time or resource-constrained deployments.

Furthermore, Gemini 2.0 Flash stands out for its minimal latency of just 4.2 seconds while achieving a solid 91.8\% accuracy. This profile makes it particularly attractive for latency-sensitive applications. In contrast, models like o1 and DeepSeek-R1 deliver exceptional accuracy scores (96.9\% and 95.8\%, respectively) but exhibit significantly longer response times (15.0 and 81.05 seconds), which may present challenges in time-critical settings. The DeepSeek-1.5B model highlights the trade-off at the other end of the spectrum, with fast inference (5.0 seconds) but substantially lower accuracy, reflecting the compromises often seen in smaller or distilled architectures. Overall, this comparison underscores a clear trade-off between accuracy and inference speed. While some models prioritize correctness at the cost of latency, others offer rapid responses with modest performance degradation. 




\section{Discussion}
\label{sec4}
In this study, we conducted a comprehensive evaluation of various LLMs to assess their mathematical reasoning capabilities across multiple benchmarks~\citep{cobbe2021gsm8k,hendrycks2020measuring,hendrycksmath2021}.Our empirical analysis reveals that o3-mini outperformed all evaluated models in terms of overall accuracy and reasoning quality across a range of mathematical domains. It demonstrated particular strength in complex and abstract areas, such as Abstract Algebra and Multistep Calculation, while maintaining relatively low latency. In comparison, DeepSeek-R1 and o1 also exhibited strong reasoning capabilities, especially in structured arithmetic domains, but their performance was slightly lower than o3-mini in both accuracy and efficiency. These results confirm the existence of a trade-off between computational efficiency and reasoning performance, which must be carefully considered when designing LLMs for downstream applications.

Furthermore, our results validate the advanced design of DeepSeek-R1, which employs GRPO—a reinforcement learning-based optimization approach—instead of conventional supervised finetuning~\citep{evstafev2025token,liu2024deepseekv2,neha2025survey}. GRPO allows the model to evaluate its outputs relative to a group of sampled peers, eliminating the need for external reward models~\citep{liu2024deepseekv2,liu2024deepseek}. This strategy achieves strong performance, particularly on GSM8K and logic-based tasks, it remains slightly behind o3-mini in accuracy and reasoning consistency across all domains. These comparisons suggest that reinforcement learning with GRPO enhances model generalization, but domain-specific prompting strategies—like those used in o3-mini—can still offer an edge in specialized reasoning tasks.

Moreover, proof-based reasoning, a crucial component of higher-level mathematics, demands meta-cognitive processes that extend beyond pattern recognition, including recursive logic application and theorem proving. The limitations observed in DeepSeek-R1 to generalize across abstract mathematical problems indicate that a more versatile reasoning approach may be needed to complement GRPO-based training. In contrast, CoT reasoning demonstrates strong performance in specific mathematical fields by leveraging step-by-step logical breakdowns to enhance problem-solving accuracy. Thus, while CoT provides a deeper, structured reasoning framework~\citep{boye2025large,li2024dotamath}, it may struggle to satisfy the generalization requirements necessary for handling diverse mathematical challenges. The trade-off between GRPO’s structured logic refinement and CoT’s sequential deductive approach suggests that a hybrid reasoning mechanism may be essential for developing a more versatile and mathematically proficient LLM.

In summary, our findings align with prior research~\citep{ahn2024large,akella2024improving,duanmeta,evstafev2025token,jiang2025deepseek,li2024evaluating,li2025system,mercer2025brief,mirzadeh2024gsm,peng2024multimath,shao2024deepseekmath,xin2024deepseek,zhong2024evaluation,zhou2024your,zhuang2024math}, further confirming that the recent invented model DeepSeek-R1 demonstrates strong mathematical reasoning compared to other LLMs. However, our research extends beyond previous work by providing a granular, field-specific analysis of LLM performance across different domains. We delve into the nuances of logical, arithmetic, and abstract reasoning capabilities, offering a holistic understanding of how different LLMs excel or struggle across mathematical disciplines. Despite its valuable contributions, our study has several limitations: 1). Limited Model Scope: We did not test other cutting-edge models, such as Qwen. 2). Restricted Benchmark Diversity. More importantly, future research should expand the scope of evaluated LLMs and incorporate additional benchmarks to gain a broader understanding of model strengths and weaknesses across varied mathematical disciplines. Notably, integrating GRPO with peer CoTs could allow models to leverage both structured logical optimization and stepwise deductive reasoning. This integration could lead to LLM swarms, where multiple reasoning frameworks synergize to enhance efficiency and accuracy. Meanwhile, our study reaffirms that distillation can significantly impair reasoning capabilities. Future efforts should explore adaptive distillation methods that preserve deep reasoning pathways while optimizing for computational efficiency.  

\section{Conclusion}
\label{sec5}
Overall, our study presents a comprehensive comparative analysis of LLM performance in mathematical reasoning, identifying key strengths and weaknesses across multiple domains. We highlight the effectiveness of GRPO in structured problem-solving, demonstrating its ability to enhance logical consistency and rule-based reasoning. However, we also uncover its limitations in handling abstract mathematical concepts, where deep symbolic manipulation and theorem-based reasoning remain challenging for GRPO-trained models. Furthermore, our findings underscore the trade-offs associated with model distillation, revealing its potential drawbacks in preserving mathematical reasoning capabilities. Beyond performance evaluation, this study paves the way for future advancements in LLM-driven mathematical intelligence by outlining potential enhancements in reasoning frameworks, including hybrid models that integrate RL with structured step-by-step inference methods. By addressing these challenges, our work contributes to the development of next-generation AI systems capable of handling complex mathematical reasoning with greater accuracy, efficiency, and adaptability.

\section{Acknowledgement}

The authors gratefully acknowledge the support of High Performance Computing Services for providing the computational resources that contributed to the results presented in this publication/report.

\bibliography{references}  
\newpage

\appendix

\section{Appendix A}

The table \ref{tab:overview_of_data} summarizes the benchmark datasets used for evaluating the mathematical reasoning abilities of LLMs. The MATH dataset consists of free-form numeric problems drawn from math competitions, covering advanced topics like algebra, geometry, number theory, and combinatorics. GSM8K focuses on grade school arithmetic through structured word problems. The MMLU subsets include four college-level and high school-level categories—College Mathematics, Abstract Algebra, Formal Logic, and High School Mathematics—each formatted as multiple-choice questions and covering domains such as discrete math, algebraic structures, and logical reasoning.

 \textbf{MATH} is derived from high-level mathematics competitions, including the AMC 10, AMC 12, and AIME~\citep{hendrycksmath2021}. These problems are inherently more complex than standard K-12 mathematics tasks, requiring the application of non-trivial problem-solving strategies, logical inference, and domain-specific heuristics rather than direct formulaic computations. Meanwhile, MATH is widely regarded as a difficult benchmark, with LLMs achieving accuracy rates between 3.0\% and 6.9\% \citep{hendrycksmath2021}. Despite these low overall scores, LLMs demonstrate some degree of mathematical competency, as evidenced by: Up to 15\% accuracy on the easiest difficulty level \citep{hendrycksmath2021}. The ability to generate step-by-step solutions that, while sometimes incorrect, remain coherent and contextually relevant. To gauge the dataset's difficulty, human performance was also considered. For instance, a computer science Ph.D. student with no particular affinity for mathematics achieved approximately 40\% accuracy. In addition, a three-time IMO (International Mathematical Olympiad) gold medalist attained 90\% accuracy. Thus, these findings indicate that MATH is challenging for both human solvers and LLMs, making it an invaluable dataset for assessing and advancing mathematical reasoning capabilities in both LLMs and human problem solvers.

 \textbf{GSM8K}, introduced in 2021, is a benchmark dataset consisting of 8,500 high-quality grade school-level math problems ~\citep{cobbe2021gsm8k}. Designed to incorporate high linguistic diversity while relying on fundamental mathematical concepts, it presents a unique challenge for state-of-the-art LLMs. Although the underlying math is relatively simple, the diverse problem formulations create significant hurdles, preventing many models from achieving consistently high accuracy. GSM8K is structured into 7,500 training problems and 1,320 testing problems, all carefully crafted by expert human problem writers. The problems primarily involve elementary arithmetic operations and typically require 2 to 8 logical steps to reach a solution. This dataset is widely used to evaluate logical reasoning and mathematical proficiency in LLMs and serves as a benchmark for various assessments, including the LLM Leaderboard. Notably, while some GSM8K problems are conceptually straightforward, they can still be challenging for even the most advanced LLMs, often exhibiting high variability in responses when the same problem is presented in slightly different ways. This highlights the ongoing difficulty of achieving robust mathematical reasoning in language models. In this study, we selected the 1.32K mathematical test set from GSM8K to systematically evaluate and compare the mathematical reasoning capabilities of each LLM.

 The \textbf{MMLU} dataset is a comprehensive benchmark comprising multiple-choice questions from a diverse range of academic disciplines. Spanning 57 distinct tasks, it covers subjects in the humanities, social sciences, hard sciences, and mathematical reasoning, reflecting the breadth of knowledge that is essential for various fields of study~\citep{hendrycks2020measuring}. MMLU’s questions were manually curated by graduate and undergraduate students from publicly available sources, including practice questions from standardized exams such as the Graduate Record Examination (GRE) and the United States Medical Licensing Examination (USMLE). Additionally, it features questions designed for undergraduate courses and readers of Oxford University Press books. Notably, its mathematical reasoning component encompasses multiple subfields, categorized into areas such as "Abstract Algebra," "College Mathematics," "Formal Logic," and "High School Mathematics." For instance, questions in the "Abstract Algebra" category originate from professional mathematical practice, while the "High School Mathematics" category includes problems akin to those found in standard high school exams. This broad and structured dataset serves as a critical benchmark for evaluating the reasoning and problem-solving capabilities of LLMs.

\begin{table}[h]
  \caption{Overview of Benchmark Data subsets Used for Evaluation}
  \label{tab:overview_of_data}
  \centering
  \resizebox{\textwidth}{!}{%
  \begin{tabular}{llcp{4cm}l}
    \toprule
    \textbf{Dataset } & \textbf{Category} & \textbf{Question Type} & \textbf{Domain} &  \textbf{Number of Problems} \\
   \midrule
     \multirow{1}{*}{MATH} 
        & Mathematics competitions & Free-form numberic & Algebra, Geometry, Number Theory, Combinatorics & 262\\
    
    \midrule
    \multirow{1}{*}{GSM8K} 
        & Grade School Math  & Word Problems & Arithmetic & 1320\\
    
     \midrule
    \multirow{4}{*}{MMLU} 
        & College Mathematics & Multiple-choice & Algebra, Calculus, Discrete Math  & 100\\
        & Abstract Algebra & Multiple-choice & Group Theory, Rings, Algebraic Structures  & 100 \\
        & Formal Logic  & Multiple-choice & Proofs, Logical Reasoning & 126\\
        & High School Mathematics  & Multiple-choice & General Mathematics & 270\\
    \bottomrule
  \end{tabular}%
  }
\end{table}

\newpage
\section{Appendix B}
\begin{figure}[ht]
     \centering
    \includegraphics[width=.9\linewidth]{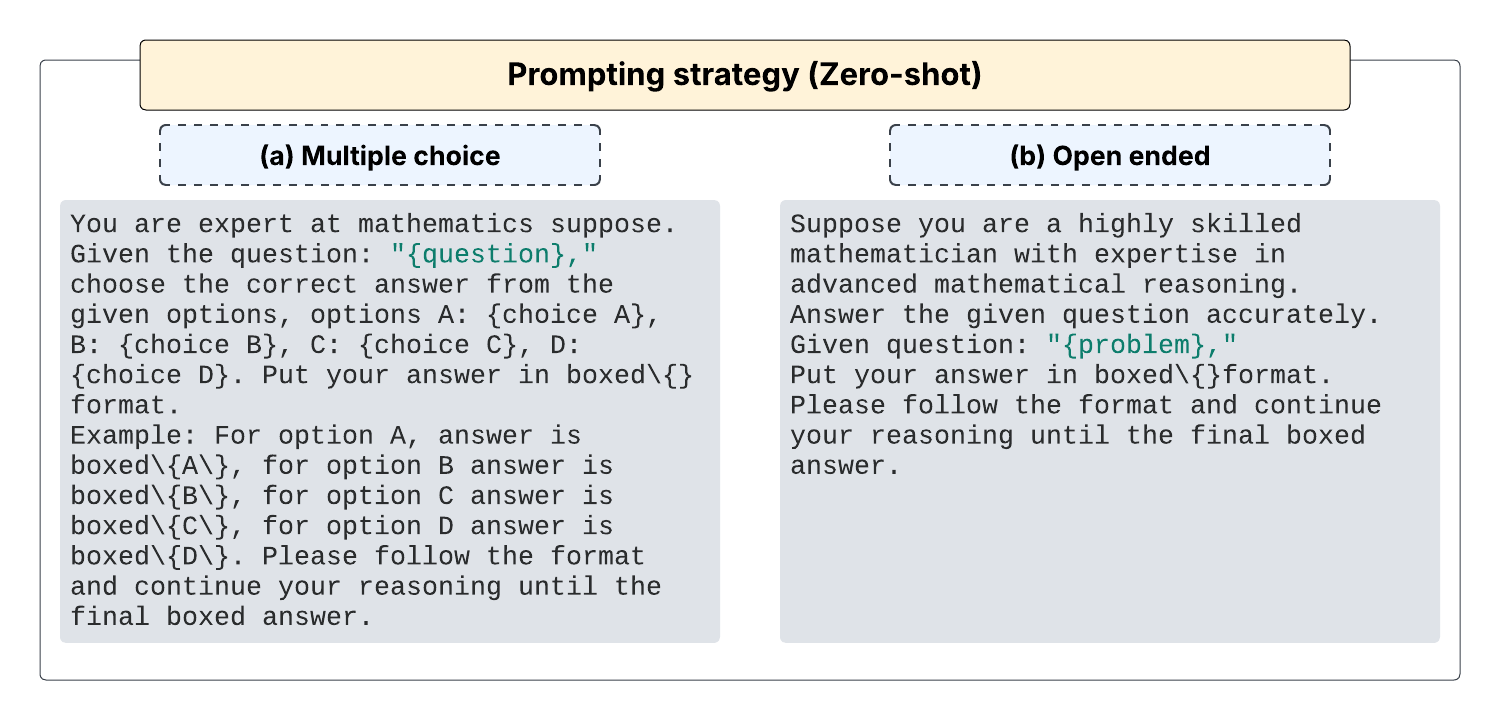} 
    \caption{Templates used to evaluate mathematical reasoning in LLMs. The multiple-choice format (a) ensures structured decision-making with predefined options, while the open-ended format (b) assesses free-form problem-solving capabilities.}
    \label{fig:math_reasoning_prompting}
\end{figure}

Each prompt as shown in Figure~\ref{fig:math_reasoning_prompting}, is explicitly instructing models to reason through the problem before selecting an answer enclosed in the \texttt{\textbackslash boxed\{\}} format.  While explicit CoT prompting was not enforced, the instruction to "continue your reasoning until the final boxed answer" implicitly encouraged stepwise reasoning, allowing us to assess the extent to which models could self-initiate logical reasoning without explicit CoT guidance~\citep{sahoo2024systematic}. Empirical observations revealed that some models naturally generated intermediate reasoning steps before selecting an answer, demonstrating emergent reasoning capabilities, while others provided the final answer directly, particularly those with weaker reasoning abilities. Variations in reasoning depth were observed across different models, reflecting differences in their internal problem-solving strategies. To further investigate the impact of explicit stepwise prompting, we conducted an ablation study by modifying the prompt to include a direct instruction. This explicit CoT version consistently led to more structured and detailed explanations across all models, confirming that models respond differently to implicit vs. explicit reasoning cues~\citep{yuan2024instance}. The comparative analysis of both prompting strategies provided valuable insights into how structured guidance influences model reasoning and accuracy.

\newpage
\section{Appendix C}

To provide a deeper understanding  of the qualitative behavior of the evaluated models, we include representative examples of both correct and incorrect responses in this section. These examples span problems from GSM8K, MMLU, and MATH benchmarks and highlight a range of model behaviors including accurate multi-step reasoning, partial computations, and common logical problems.

\begin{figure}[ht]
     \centering
    \includegraphics[width=\linewidth]{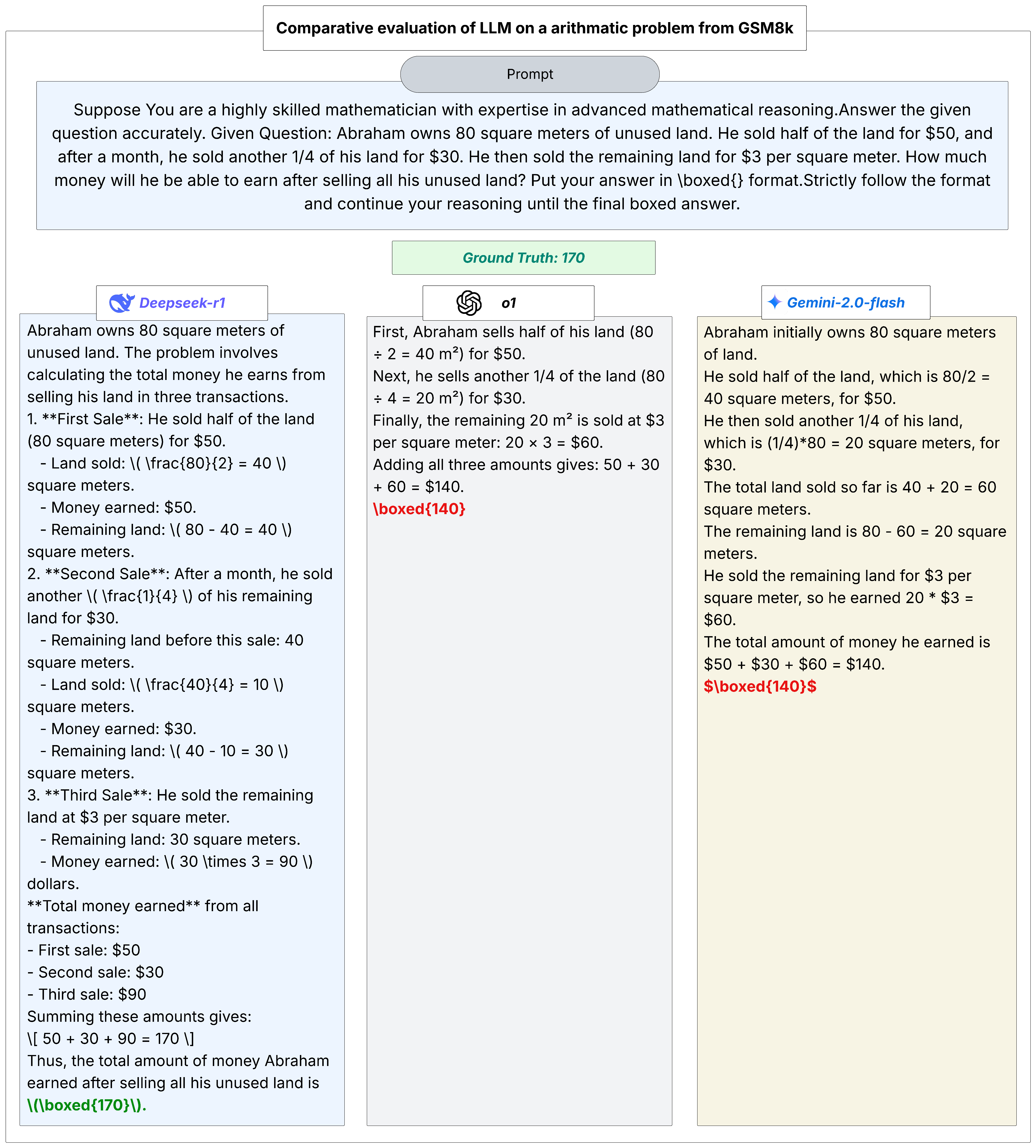} 
    \caption{Comparative evaluation of different LLMs on an Arithmetic problem from the GSM8K benchmark. The figure showcases step-by-step reasoning from three different models and highlights correctness. The question is shown in red, while the extracted answers are boxed. Green-highlighted sections indicate correct responses, whereas red-highlighted sections denote incorrect outputs. This comparison illustrates differences in mathematical reasoning and accuracy across models}
    \label{fig:model_evaluation_result_gsm8k}
\end{figure}

\newpage

\begin{figure}[ht]
     \centering
    \includegraphics[width=\linewidth]{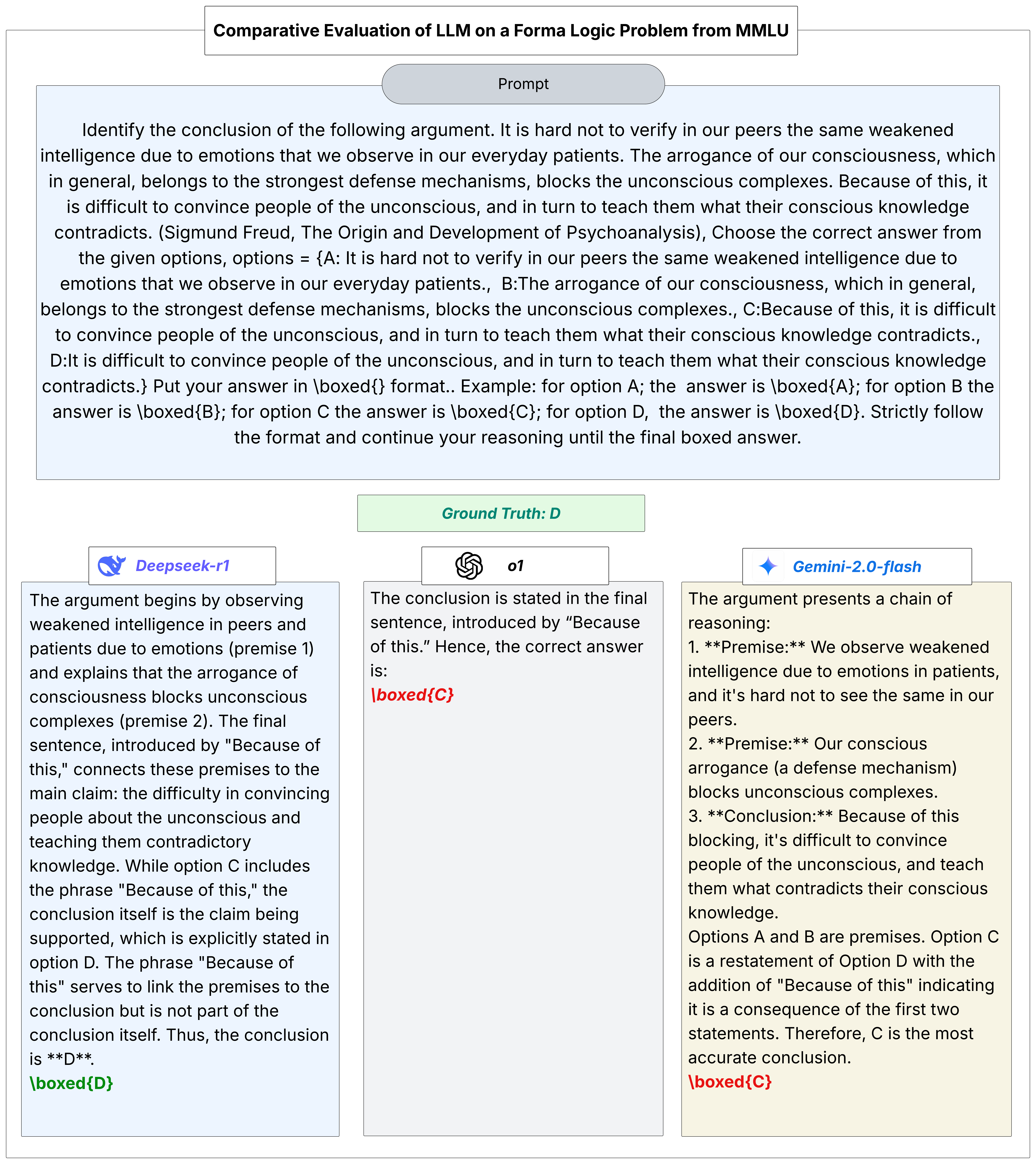} 
    \caption{Comparative evaluation of different LLMs on a Formal Logic problem from the  MMLU benchmark. The figure showcases step-by-step reasoning from three different models and highlights correctness. The question is shown in red, while the extracted answers are boxed. Green-highlighted sections indicate correct responses, whereas red-highlighted sections denote incorrect outputs. This comparison illustrates differences in mathematical reasoning and accuracy across models.}
    \label{fig:model_evaluation_result_formal_logic_mmlu}
\end{figure}

\newpage

\begin{figure}[ht]
     \centering
    \includegraphics[width=\linewidth]{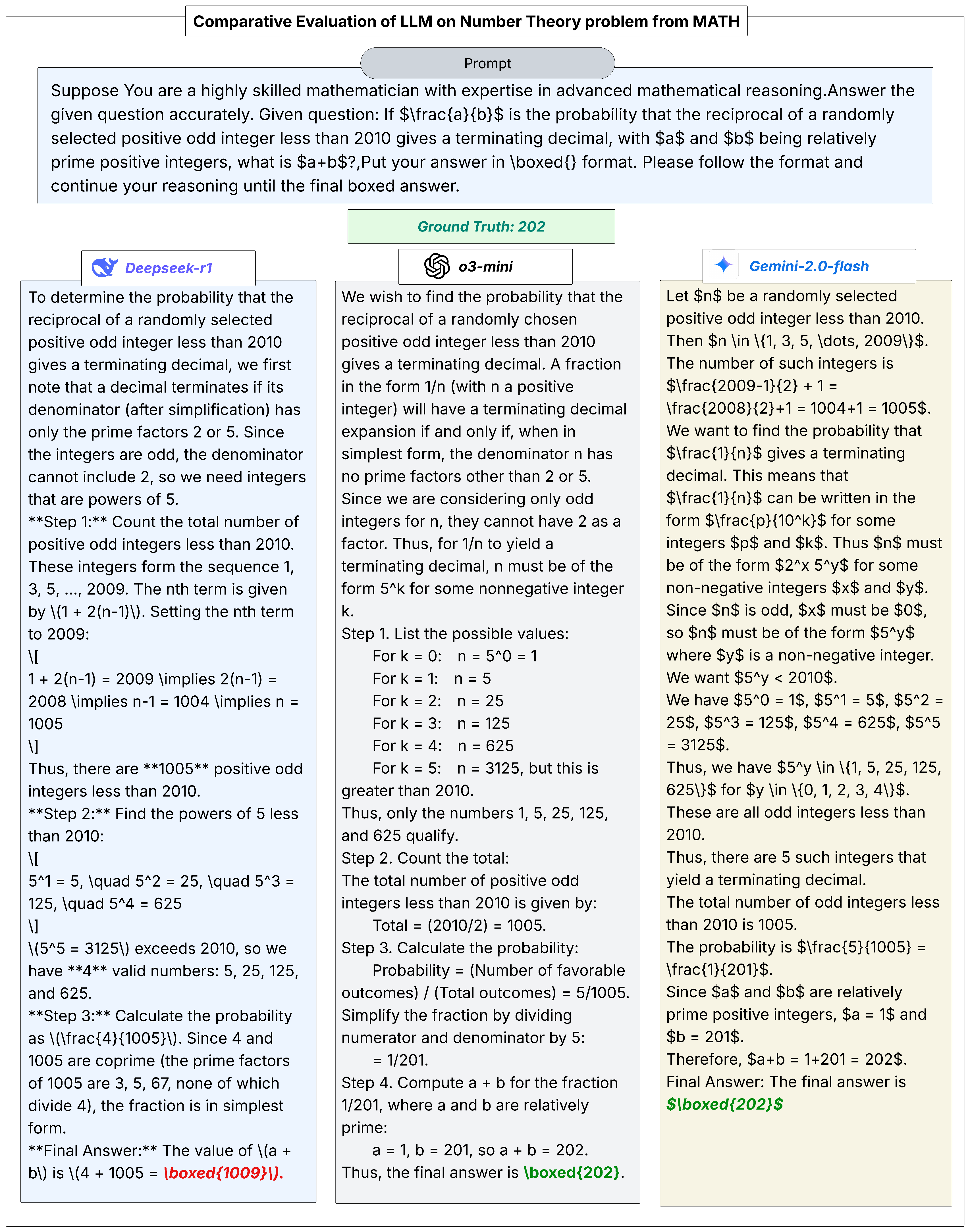} 
    \caption{Comparative evaluation of different LLMs on a Geometric problem from the MATH benchmark. The figure showcases step-by-step reasoning from three different models and highlights correctness. The question is shown in red, while the extracted answers are boxed. Green-highlighted sections indicate correct responses, whereas red-highlighted sections denote incorrect outputs. This comparison illustrates differences in mathematical reasoning and accuracy across models.}
    \label{fig:model_evaluation_example_math}
\end{figure}

\newpage

\section{Appendix D}

\begin{figure}[h]
    \centering
    \includegraphics[width=0.90\linewidth]{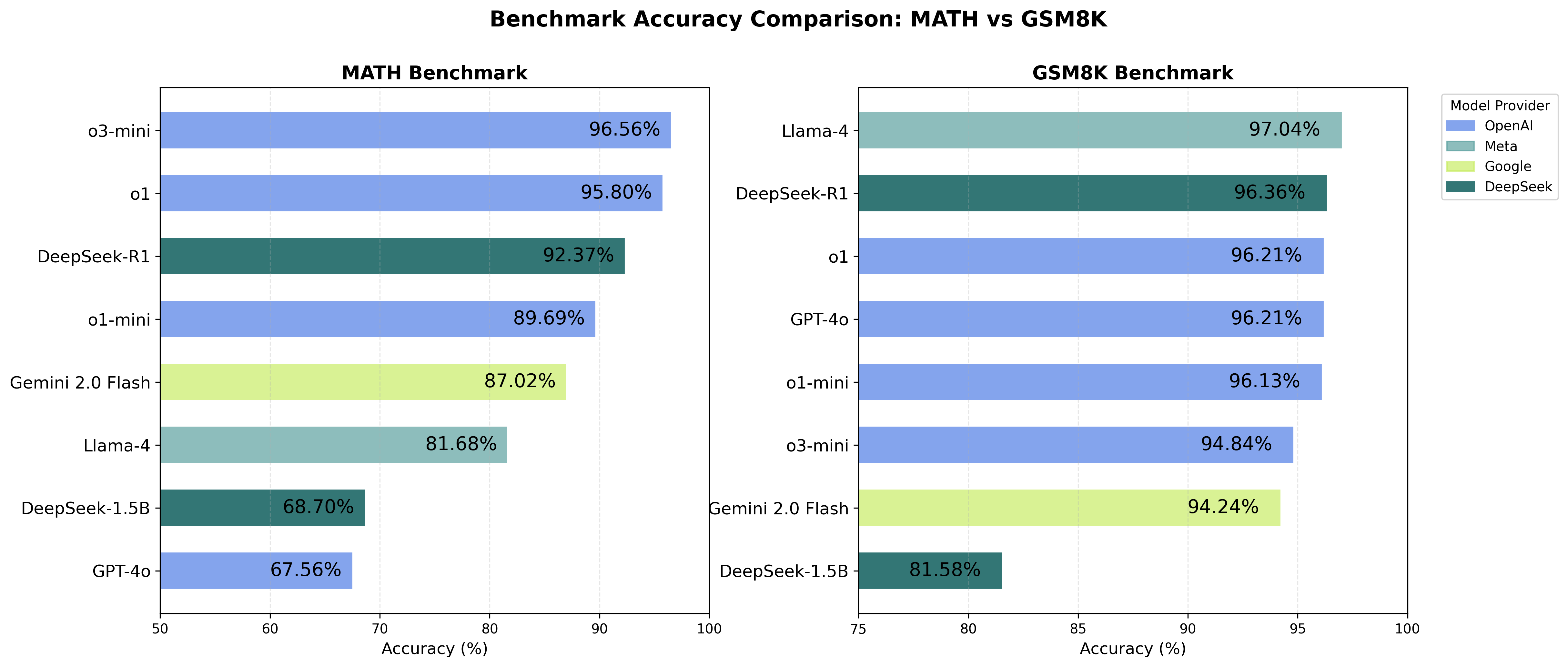} 
    \caption{Performance comparison of various LLMs in the GSM8K and MATH benchmark. The accuracy of each model is displayed above the corresponding bar, highlighting differences in mathematical problem-solving capabilities across model publishers.}
    \label{fig:res_fig_GSM8K_MATH}
\end{figure}

\begin{figure}[h]
    \centering
    \includegraphics[width=0.90\linewidth]{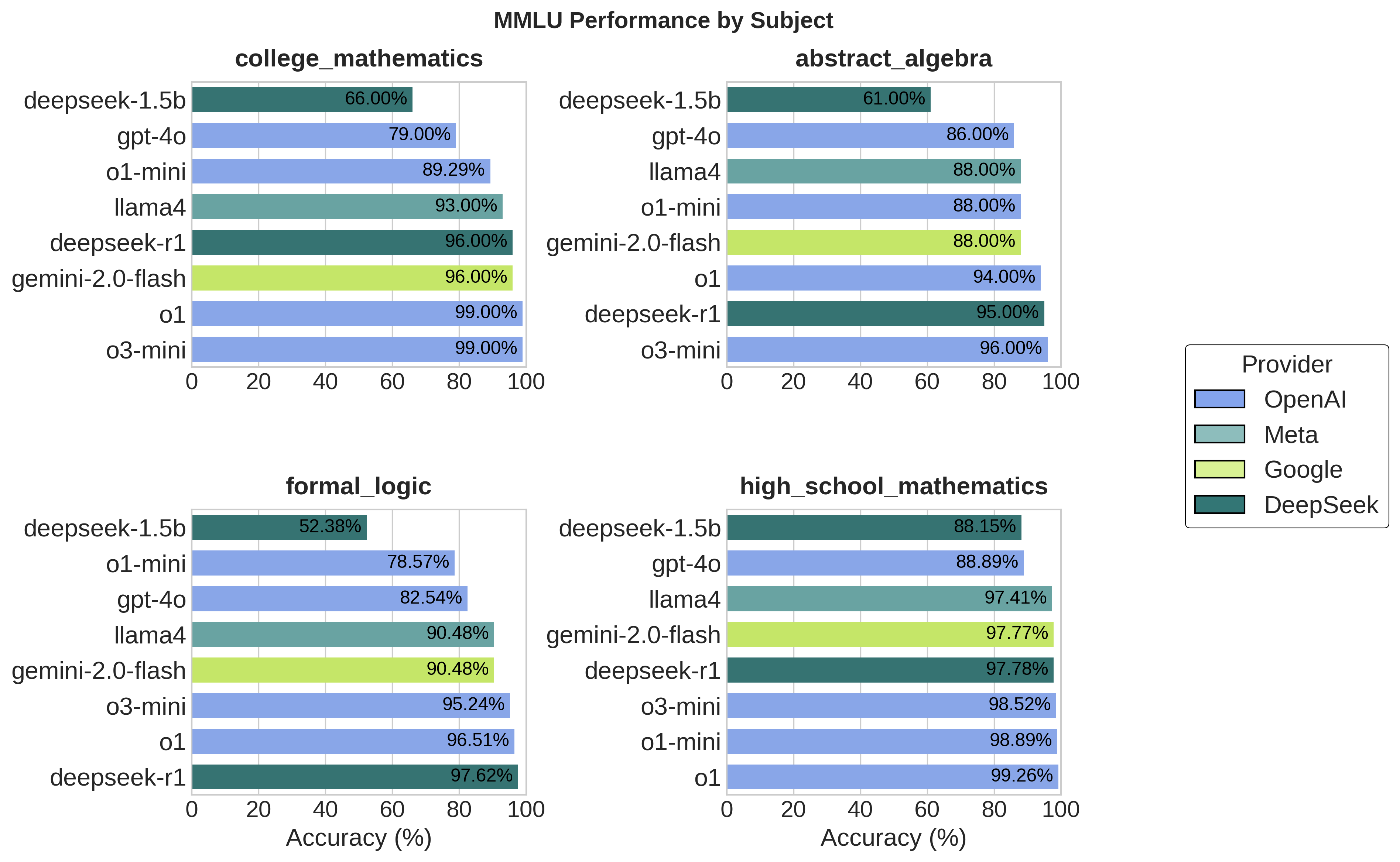} 
    \caption{Accuracy comparison of large language models (LLMs) across four mathematical MMLU sub-domains: \textit{College Mathematics}, \textit{Abstract Algebra}, \textit{Formal Logic}, and \textit{High School Mathematics}. This visualization highlights performance variance across both model architectures and mathematical domains, showcasing how some models (e.g., o3-mini, o1) maintain consistently high scores across all categories, while others (e.g., DeepSeek-1.5B) struggle in more abstract reasoning tasks like Formal Logic and Algebra.}
    \label{fig:mmlu_each_section_eval}
\end{figure}

\begin{figure}[h]
    \centering
    \includegraphics[width=0.95\linewidth]{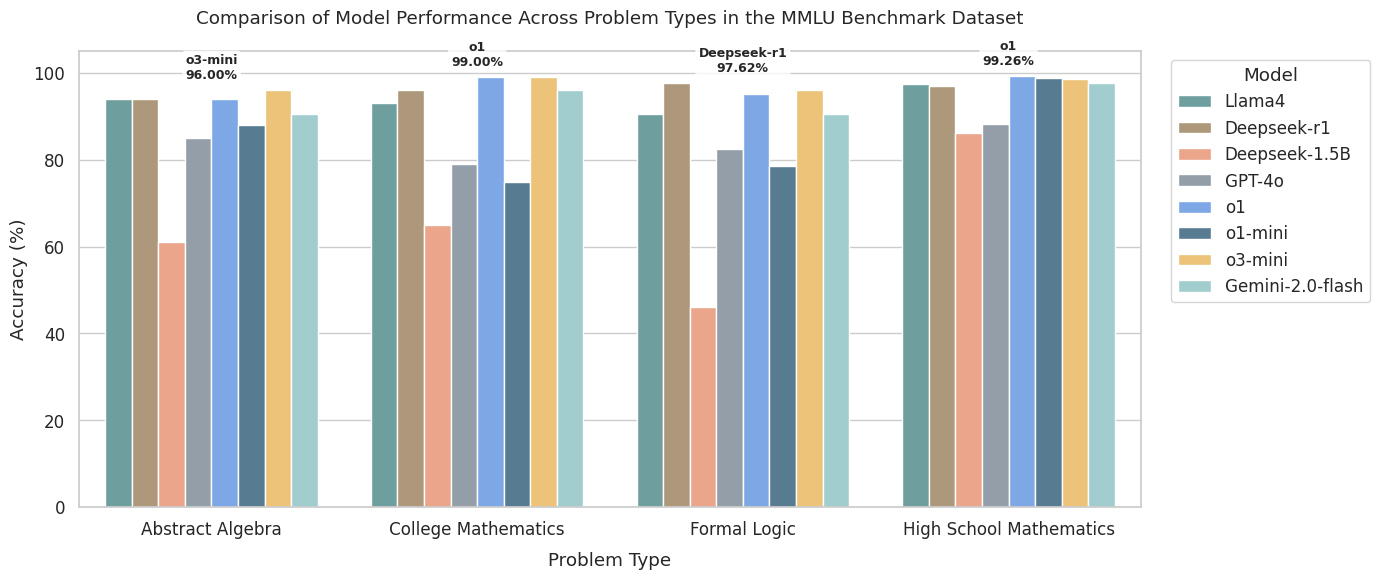} 
    \caption{Comparison of LLM models across different problem types in the MMLU benchmark dataset.}
    \label{fig:res_fig_MMLU}
\end{figure}


\newpage

\end{document}